\documentclass[conference]{IEEEtran}
\IEEEoverridecommandlockouts
\pdfoutput=1 
\usepackage{cite}
\usepackage{amsmath,amssymb,amsfonts}
\usepackage{algorithmic}
\usepackage{graphicx}
\usepackage{textcomp}
\usepackage{xcolor}
\usepackage{multicol}
\usepackage{multirow}
\def\BibTeX{{\rm B\kern-.05em{\sc i\kern-.025em b}\kern-.08em
    T\kern-.1667em\lower.7ex\hbox{E}\kern-.125emX}}

\begin{document}

\title{A CNN Approach to Automated Detection and Classification of Brain Tumors}

\author{
    \IEEEauthorblockN{
        Md. Zahid Hasan\textsuperscript{1}, 
        Abdullah Tamim\textsuperscript{2}, 
        D.M. Asadujjaman\textsuperscript{3}, 
        Md. Mahfujur Rahman\textsuperscript{4}\\ 
        Md. Abu Ahnaf Mollick\textsuperscript{5}, 
        Nosin Anjum Dristi\textsuperscript{6}, 
       Abdullah - Al - Noman\textsuperscript{7}
    }
 \IEEEauthorblockA{
        \textsuperscript{1, 2, 7}Dept. of Computer Science \& Engineering, University of Rajshahi, Bangladesh \\
        \textsuperscript{3}Dept. of Computer Science \& Engineering, Khulna University of Engineering \& Technology, Khulna, Bangladesh \\
        \textsuperscript{4}Dept. of Computer Science \& Engineering, Rajshahi University of Engineering \& Technology, Rajshahi, Bangladesh \\
        \textsuperscript{2, 3, 4, 5, 6 }Dept. of Computer Science \& Engineering, Varendra University, Rajshahi, Bangladesh \\
    }
    \IEEEauthorblockA{
        jahidhasan110410@gmail.com, 
        abtamim131415@gmail.com, 
        asadujjaman@vu.edu.bd, 
        mahfujur@vu.edu.bd,\\ 
        mollickavoy@gmail.com, 
        nosindristi040@gmail.com, 
        abnoman23rd@gmail.com
    }
}


\maketitle

\begin{abstract}
Brain tumors require an assessment to ensure timely diagnosis and effective patient treatment. Morphological factors such as size, location, texture, and variable appearance complicate tumor inspection. Medical imaging presents challenges, including noise and incomplete images. This research article presents a methodology for processing Magnetic Resonance Imaging (MRI) data, encompassing techniques for image classification and denoising. The effective use of MRI images allows medical professionals to detect brain disorders, including tumors. This research aims to categorize healthy brain tissue and brain tumors by analyzing the provided MRI data. Unlike alternative methods like Computed Tomography (CT), MRI technology offers a more detailed representation of internal anatomical components, making it a suitable option for studying data related to brain tumors. The MRI picture is first subjected to a denoising technique utilizing an Anisotropic diffusion filter. The dataset utilized for the model's creation is a publicly accessible and validated Brain Tumour Classification (MRI) database, comprising 3,264 brain MRI scans. SMOTE was employed for data augmentation and dataset balancing. Convolutional Neural Networks(CNN) such as ResNet152V2, VGG, ViT, and EfficientNet were employed for the classification procedure. EfficientNet attained an accuracy of 98\%, the highest recorded.
\end{abstract}

\begin{IEEEkeywords}
MRI, EfficientNet, Brain Tumor, SMOTE, CNN
\end{IEEEkeywords}

\section{Introduction}

The brain and the spinal cord are combinedly known as the Central Nervous System, which is crucial for the control of numerous cellular functions. The functions include organization, analysis, decision-making, directive issuance, and information integration \cite{b1}. The human brain exhibits extraordinary complexity owing to its distinctive physical architecture. Conditions such as brain tumors, infections, migraines, and strokes are a small subset of central nervous system (CNS) disorders that present considerable challenges in the development, diagnosis, and assessment of successful treatment strategies, as noted in \cite{b3}. A significant difficulty for radiologists and neuropathologists is the early detection of brain tumors, which result from the aberrant proliferation of brain cells. The detection of cerebral malignancies by magnetic resonance imaging (MRI) is an intricate manual procedure prone to inaccuracies. The abnormal proliferation of nerve cells, resulting in the creation of a mass, is termed a brain tumor. There are approximately 130 unique types of tumors that can arise in the brain and the central nervous system, encompassing both benign and malignant forms. The prevalence of various tumors differs, with some being exceptionally rare and others often observed \cite{b4}. Till date approximately 700,000 individuals in the United States are diagnosed with primary brain tumors. According to research cited in \cite{b5}, patients between the ages of 55 and 64 had a 46.1\% one-year survival rate, whereas patients between the ages of 65 and 74 had a 29.3\% survival rate. Image segmentation is a technique used to divide a image into several segments, frequently employed in the medical imaging domain. The visual representation of a picture can be enhanced by extraction for analytical purposes. This transpires when the image is partitioned into multiple distinct segments. The scientific examination of images predominates in the field of medical diagnostics. The existence of nuanced differences, specific types of noise, and the absence of evidence concerning impediments in medical imaging complicates the resolution of this issue. However, magnetic resonance imaging (MRI) is more advantageous than autonomous computed tomography (CT) equipment. It emits no radiation and hence has no adverse effects on the human body. The fundamental elements are the magnetic field and radio waves. It is a widely utilized non-invasive imaging modality that offers accurate distinction between tissues. The capability of MRI to normalize commonly affected tissue enhances the imaging of structures of interest in human brain tumors. Researchers have lately encountered a significant obstacle in the manual segmentation of brain MRI images \cite{b7}. Image segmentation has often been employed to identify brain tumors. Diverse methodologies necessitate a patient-specific training dataset to perform tailored MRI tumor imaging investigations.These types of datasets intensify the difficulties for specialists. And these solutions typically depend on alternate imaging modalities, such as T1-weighted contrast-enhanced images. Various research also shows the use of Synthetic Minority Oversampling Technique (SMOTE) for data augmentation and balancing the dataset \cite{b20}.

The rapid growth and unfavorable prognosis of brain tumors, particularly glioblastomas, represent a significant health threat. Despite advancements in MRI and CT imaging, numerous cancers are diagnosed late owing to ambiguous symptoms. The location of the tumor and the blood-brain barrier complicates treatment. Glioblastomas are the most common and malignant form of brain cancer, with a median survival of little over one year in advanced stages \cite{b26}. Advancements in neuroimaging, such as functional MRI and Positron Emission Tomography (PET) scans, enhance tumor detection \cite{b24}. Novel immunotherapy and targeted therapies are under investigation \cite{b25}. The classification of brain tumors is essential for resolving these concerns. Histological and molecular classifications assist clinicians in predicting tumor behavior and tailoring treatment options, thereby enhancing outcomes and minimizing side effects. Brain Tumor classification aids in identifying biomarkers for focused therapy and improved prognostic outcomes. AI methods such as Deep learning and Machine Learning enhance classification precision, hence augmenting diagnosis and treatment \cite{b21, b22, b23}.

This research categorizes four tumor types: Glioma, No Tumor, Meningioma Tumor, and Pituitary Tumor, utilizing deep learning techniques to enhance classification precision. The models evaluate MRI scan data to enhance the reliability and speed of brain tumor classification, hence aiding in diagnosis and treatment planning.

\section{Related Works}
We have assessed multiple previous research efforts related to machine learning-based supervised, semi supervised, and unsupervised algorithms relevant to time series analysis. We conducted a thorough analysis to identify the shortcomings of the current system. This research improved the classification framework we developed for brain cancers.
\begin{itemize}
\item Pendela Kanchanamala et al. \cite{b9} utilized MRI to developed an optimization-enhanced hybrid deep learning model for classification and detection of brain tumors.

\item Emrah Irmak \cite{b10} attained an accuracy of 92.66\% utilizing a bespoke CNN model for classifying normal, meningioma, pituitary, glioma and metastatic brain tumors.

\item For classification of brain cancer Ayadi et al. \cite{b11} proposed a CNN-based computer assisted diagnosis (CAD) method. Three separate datasets were used to conducted the experiment using 18-weighted layered CNN model. Where they achieved 94.74\% classification accuracy for brain tumor type classification and for tumor grading, they achieved 90.35\%. 

\item Khan et al. \cite{b12} (2020) introduced a deep learning approach for the classification of brain cancers as malignant or benign, utilizing 253 genuine brain MRI scans supplemented with data augmentation techniques. Edge detection was employed to define the region of interest in the MRI image before feature extraction using a basic CNN model. The achieved categorization accuracy was 89\%.

\item The potential of deep learning techniques for glioma classification by MR imaging is examined by Banerjee et al. \cite{b13}. For 2D images the researchers assessed the effectiveness of transfer learning employing VGGNet and ResNEt architectures, attaining accuracies of 84\% and 90\%. 

\item In a distinct study \cite{b14}, researchers proposed two methodologies for glioma grading, which involved segmentation utilizing a customized U-Net model. A regional convolutional neural network (R-CNN) was employed for the classification job in each two-dimensional image slice of the MRIs. Their proposed 2D Mask R-CNN achieved an accuracy of 96\%.Data augmentation enhanced the outcomes, as evidenced by the classification efficacy of the 2D model.

\item In the research conducted by A. M. Dikande Simo, two models were proposed, trained utilizing the Brain Tumor MRI Dataset \cite{b27}. Four optimizers were evaluated across three classification tasks, with Adam demonstrating superior performance in differentiating tumor from non-tumor brains.
Where they got 100\% training accuracy and 98\% validation and test accuracy.

\item In the research of \cite{b28} A. Nag et al. introduces TumorGANet, a sophisticated model that integrates ResNet50 and Generative Adversarial Networks (GANs) for the classification of brain tumors. The model demonstrates exceptional accuracy of 99.53\% and achieves precision and recall rates of 100\%, supported by Explainable AI methodologies such as LIME. Nevertheless, the dependence on a particular dataset and the restricted examination of real-world clinical variability may limit its generalizability.

\item In the research conducted by  A. Rath, B. S. P. Mishra, and D. K. Bagal,\cite{b29} they uesd pretrained ResNet50 model to improve  accuracy and efficency utilizing a balanced dataset of 2,577 MRI images having binary class of tumors and healthy instances of patients.

\end{itemize}
A particular application, certain machine learning models demonstrate superior efficacy compared to others. However, the effectiveness of these models in classifying cardiovascular illnesses has not yet reached parity. Further advancement is required to enhance the existing work. Our motivation is to contribute to the field using deep learning methods to improve the performance and efficiency of detection and classification.

\section{Methodology}
\begin{figure*}[htbp]
\centerline{\includegraphics[width=1\textwidth, height=0.3\textwidth]{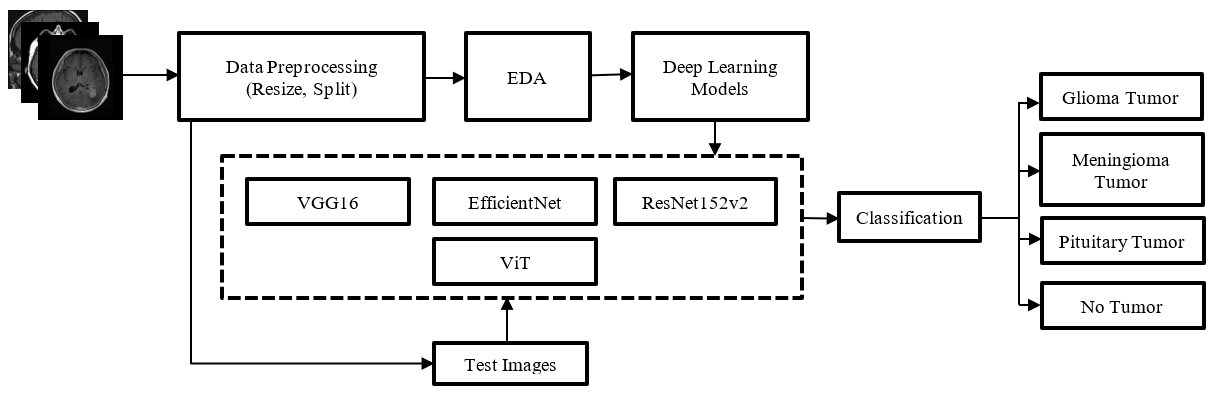}}
\caption{The proposed methodology of the system.}
\label{fig1}
\end{figure*}

This study involves a systematic approach to analyzing the Brain
tumor classification (MRI) dataset to perform the most accurate classification of brain tumors using various deep-learning models. The process is outlined in the accompanying structural outline, Fig. \ref{fig1} illustrates the key steps and stages of the analysis.

Fig. \ref{fig1} illustrates the process of deep learning by using four neural networks, namely ResNet152V2, VGG, ViT, and EfficientNet model. Every model separately process the supplied data and produce predictions. The models are subsequently compared to determine the most accurate prediction.

Utilizing a deep learning methodology, models are developed to calculate accuracy and assess predictions of different classes of brain tumors. The models can incorporate the distinctive characteristics and performance measurements that are pertinent to each position by classifying the tumors. In addition, the evaluation of the models involves the analysis of the f1 score, confusion matrix, and Receiver Operating Characteristic (ROC) curves.
\subsection{Dataset and Experiment}
The pipeline begins with feeding the image data into the system. These input images can be part of any image classification dataset. This study's dataset comprises four classes of various types of tumors, which are used to train and evaluate the models. The classes are Glioma Tumor, No Tumor, Meningioma Tumor, and Pituitary Tumor.
\begin{table}[htbp]
\caption{Details of Dataset}
\centering
\begin{tabular}{c c c c c}
\hline
\textbf{Dataset} & \textbf{Class} & \textbf{MRI/ Class} & \multicolumn{2}{ c }{\textbf{Dataset}} \\
\cline{4-5}
 &  &  & \textbf{Train} & \textbf{Test} \\
\hline

\multirow{4}{*}{Dataset\cite{b19}}

  & No Tumor     & 938 & 833 & 105 \\
  & Glioma Tumor         & 941 & 841 & 100 \\
  & Meningioma Tumor & 929 & 814 & 115 \\
  & Pituitary Tumor & 923 & 849 & 74 \\
\hline
\multirow{4}{*}{Balanced Dataset}

  & No Tumor     & 946 & 841 & 105 \\
  & Glioma Tumor         & 941 & 841 & 100 \\
  & Meningioma Tumor & 956 & 841 & 115 \\
  & Pituitary Tumor & 915 & 841 & 74 \\
\hline
\end{tabular}
\label{tabv1}
\end{table}
\subsubsection{Exploratory Data Analysis}
\begin{figure}[htbp]
\centerline{\includegraphics[width=0.5\textwidth]{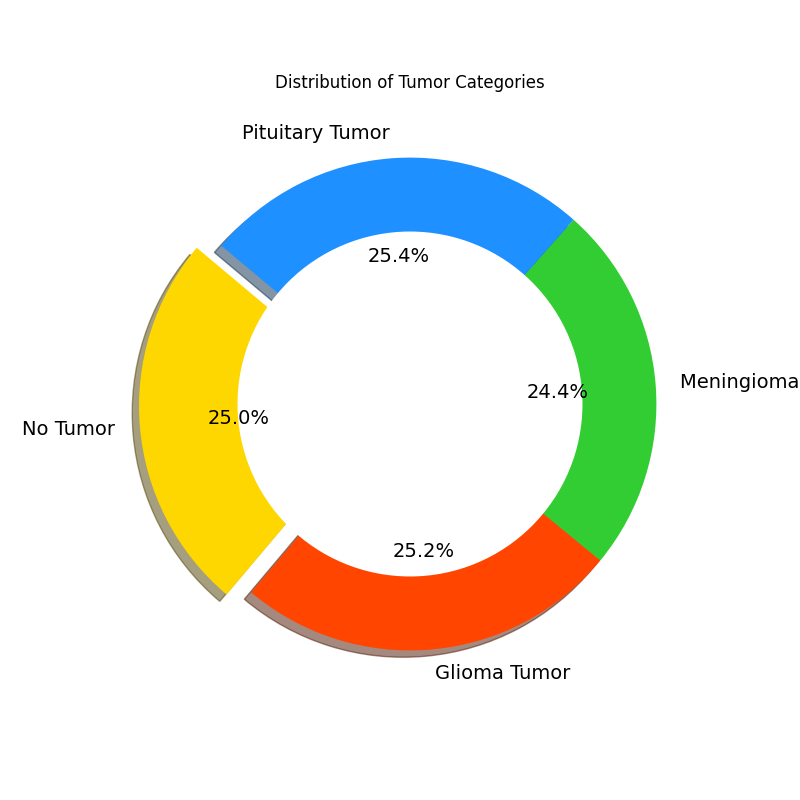}}
\caption{The distribution of the original training dataset.}
\label{fig2}
\end{figure}

Exploratory Data Analysis (EDA) is applied to summarize the main features of the data, discover patterns, and identify any potential issues such as faulty images or outliers. Fig. \ref{fig2} shows the distribution of the training dataset across four different classes of brain tumors. Glioma Tumor forms 25.20\% of the dataset used for training. Furthermore, meningioma tumors constitute 24.39\% of the training sample. The Pituitary Tumor class covers an additional 25.44\% of the training data. The No Tumor category comprises the smallest fraction, accounting for 24.96\% of the training sample.

\begin{figure}[htbp]
\centerline{\includegraphics[width=0.5\textwidth]{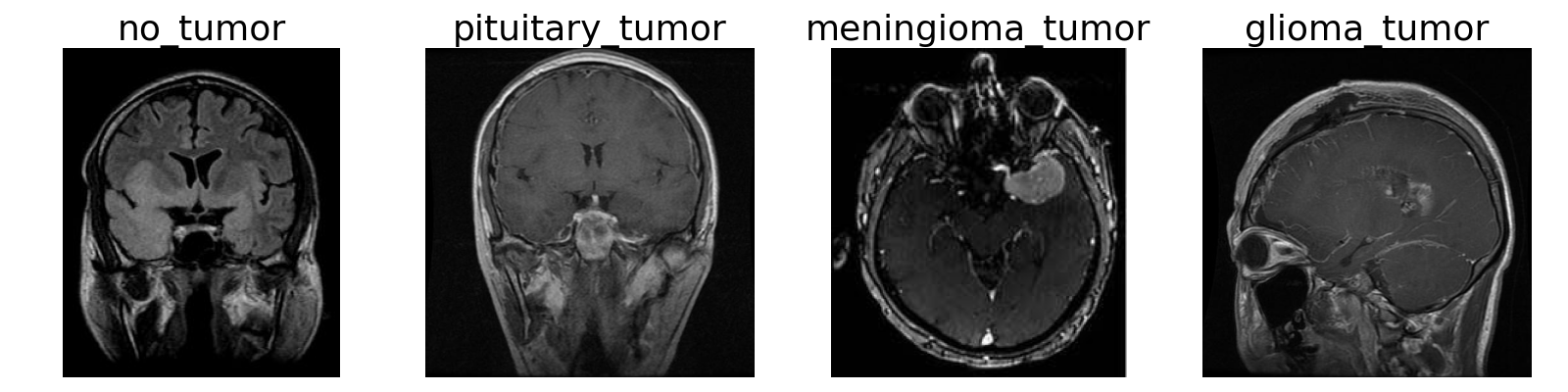}}
\caption{Train image data from data augmentation.}
\label{fig3}
\end{figure}

Fig. \ref{fig3} illustrates a grid consisting of four MRI brain scans, each annotated with a specific category of brain tumor: glioma, pituitary, or meningioma. The images are monochromatic and depict cross-sections of the human brain acquired from various perspectives and planes. Multiple scans reveal the presence of glioma and pituitary tumors, as well as the meningioma tumor. The aforementioned images have significance in the fields of medical diagnosis, treatment planning, and research about brain tumors.

Table \ref{tabv1} shows that the original training dataset maintains an equitable distribution of around 850 photos for each type of tumor (meningioma, glioma, and pituitary), but has a smaller sample size (around 800) for the category of "Meningioma Tumor". The test set shows a comparable trend, with approximately 100 photos allocated to each tumor group and a somewhat smaller number (roughly 75) for the "Pituitary Tumor" category.

\subsubsection{Dataset Preprocessing}
A ratio of 80:20 is used to divide the training and test datasets. The training dataset is used to train the model, whereas the test dataset is used to evaluate the model's performance during training. 
Neural networks require inputs of a fixed size, requiring dimensional image adjustment. The size is reliant upon the architecture of the model being applied. It requires images with exactly 224x224 pixel size. Proper scaling ensures uniformity across the dataset. Also a well-balanced dataset is necessary for the successful training of a machine learning model, taking into account the presence of a somewhat under-represented "no tumor" class, especially in the training set. Therefore, the Synthetic Minority Oversampling Technique (SMOTE) is employed for data augmentation and to balance the dataset by generating synthetic samples. This approach effectively addresses class imbalance, enhancing the performance of machine learning models.
Label encoding is a technique used to convert string labels into numerical values. This includes one-hot encoding, which converts them into a specific binary vector representation.
In addition, shuffling is implemented to ensure the random assortment of data, thereby improving the learning skills of the model and reducing the detection of organized patterns in the dataset. Before partitioning the data into training and test sets, it is imperative to restructure it.
\subsubsection{Hyperparameters}
Hyperparameters in CNNs, including layer number, filter size, learning rate, batch size, and epochs influence a model's performance. The learning rate regulates adjustments. The hyperparameters of the experiment are presented in the table.

\begin{table}[h!]
\caption{Models parameters for input and classification stage}
\centering
\begin{tabular}{ l l l }
\hline
\textbf{Parameters}          & \textbf{CNN Models} & \textbf{ViT} \\ \hline
Input Shape                  & 224 $\times$ 224    & 72 $\times$ 72\\           
No. of epochs                & 100                 & 100    \\                  
Batch Size                   & 32                  & 16    \\                     
Activation Function          & Softmax             & Softmax                      \\
Learning Rate                & 0.0001              & 0.001                        \\ 
Patch Size                   & -                   & 3                            \\ 
No. of patches               & -                   & (72/3) $\times$ 2            \\ 
Transformation Layers        & -                   & 8                            \\ \hline
\end{tabular}

\label{tab}
\end{table}

\subsubsection{Experimental Setup}
This experiment uses 8GB Nvidia GeForce RTX 4060 and 32 GB RAM. We used NumPy, Sklearn, Matplotlib, Seaborn, Python 3.10, and TensorFlow to create pre-trained networks.

\section{Result and Discussion}
Deep-learning models are trained using preprocessed data and selected classes. This work employs ViT, ResNet152V2, VGG16, and EfficientNet. The confusion matrix and ROC curves are created to get optimal results and provide a visual representation. Next, the models are evaluated using several metrics.

\begin{figure*}[htbp]
    \begin{minipage}[t]{0.48\textwidth}
        \centering
        \includegraphics[width=\textwidth, height=1.1\textwidth]{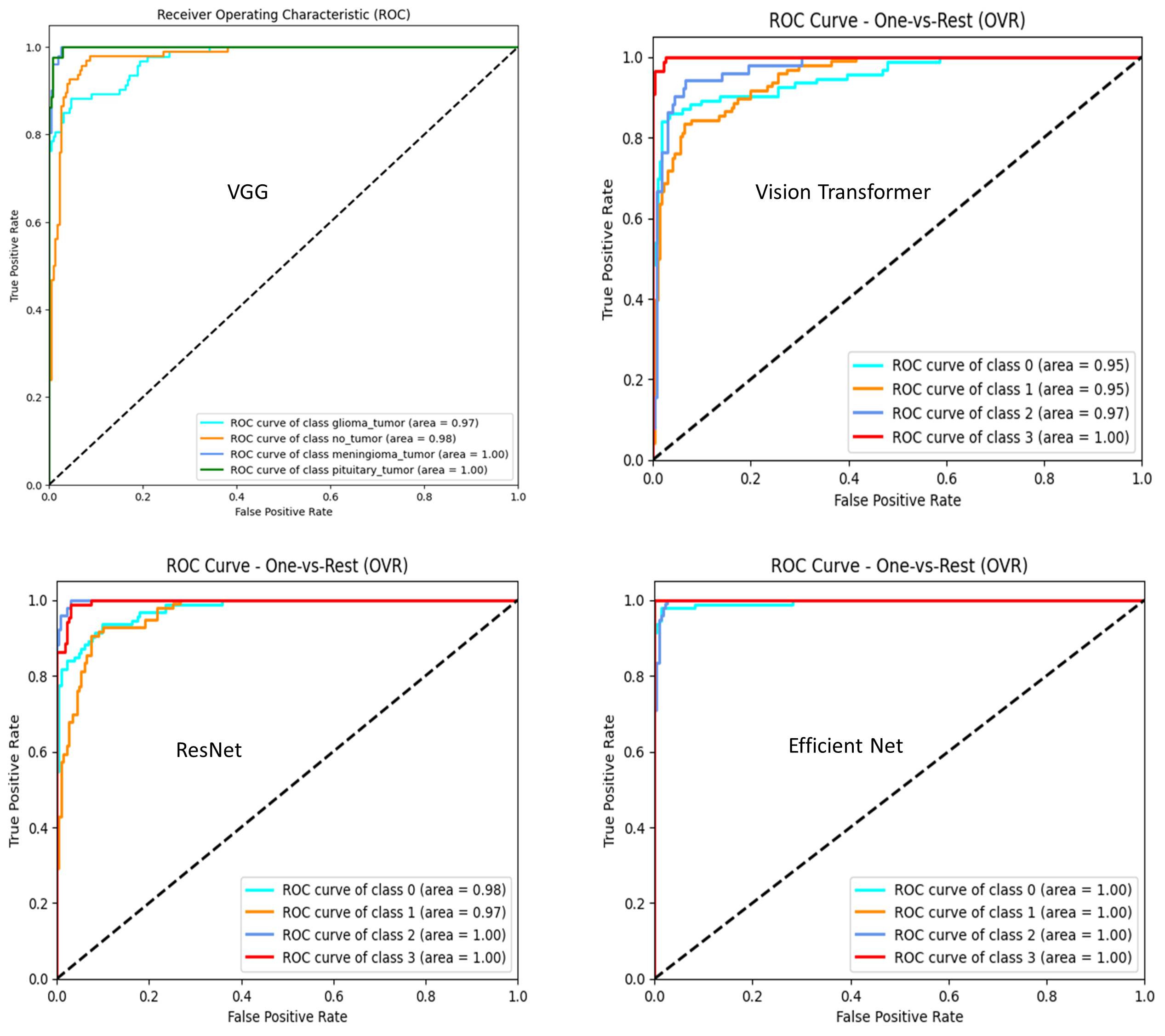}
        \caption{ROC curves for all models.}
        \label{fig5}
    \end{minipage}
    \hfill
    \begin{minipage}[t]{0.48\textwidth}
        \centering
        \includegraphics[width=\textwidth, height=1.1\textwidth]{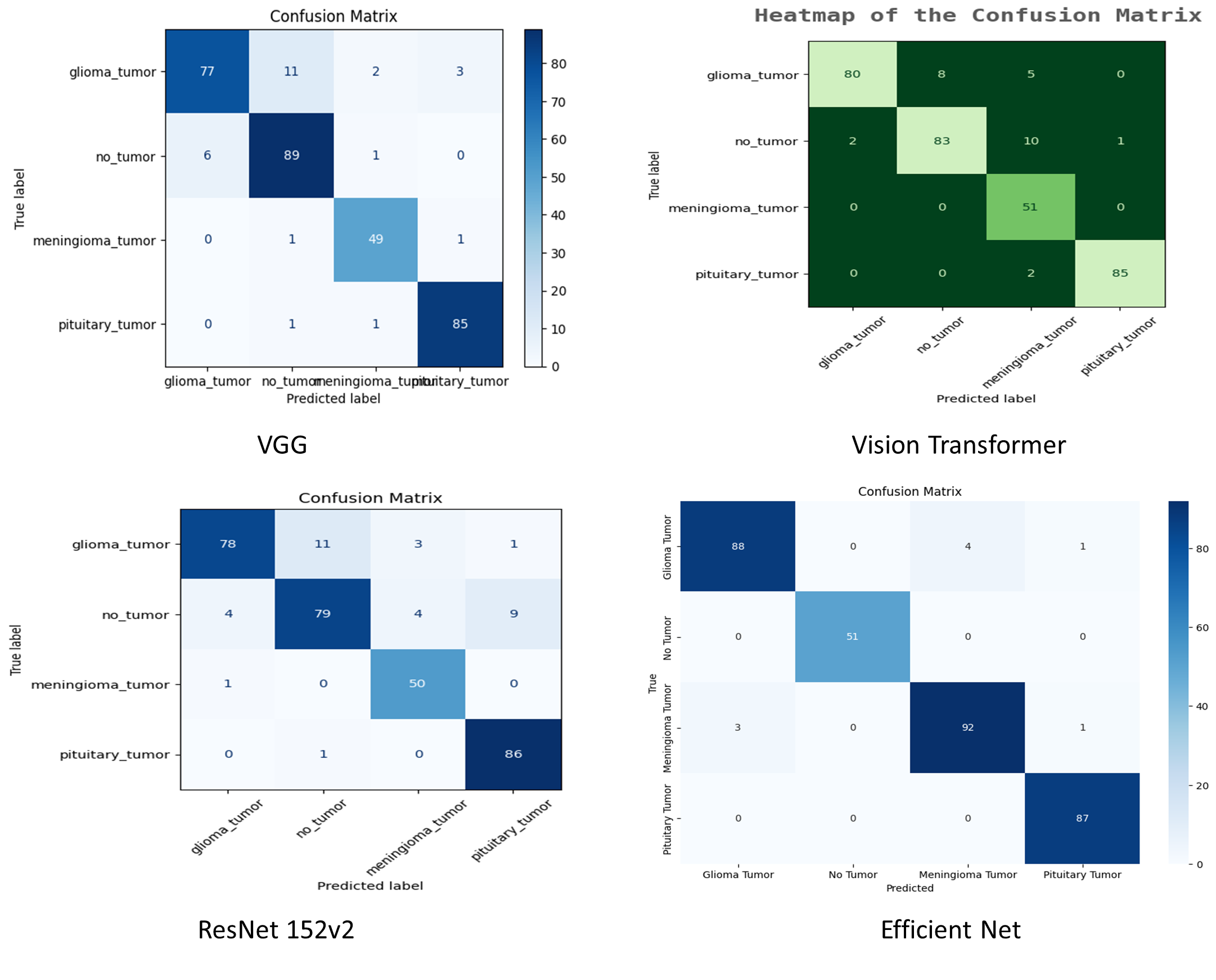}
        \caption{Confusion matrix for all models.}
        \label{fig6}
    \end{minipage}
\end{figure*}

\begin{table*}[htbp]
\caption{Precision, Recall, and F1 scores for all models}
\centering
\begin{tabular}{ c ccc| ccc| ccc| ccc }
\hline
\textbf{Classes} & \multicolumn{3}{ c| }{\textbf{VGG 16}} & \multicolumn{3}{ c| }{\textbf{EfficientNet}} & \multicolumn{3}{ c| }{\textbf{ResNet152V2}} & \multicolumn{3}{ c }{\textbf{ViT}} \\
\cline{2-13}
 & \textbf{Precision} & \textbf{Recall} & \textbf{F1} & \textbf{Precision} & \textbf{Recall} & \textbf{F1} & \textbf{Precision} & \textbf{Recall} & \textbf{F1} & \textbf{Precision} & \textbf{Recall} & \textbf{F1} \\
\hline
Glioma tumor     & 0.92 & 0.89 & 0.91 & 1.00 & 0.97 & 0.98 & 0.97 & 0.81 & 0.88 & 0.90 & 0.92 & 0.91 \\
No tumor         & 0.94 & 0.89 & 0.91 & 0.96 & 1.00 & 0.98 & 0.85 & 0.90 & 0.87 & 0.92 & 0.81 & 0.86 \\
Meningioma tumor & 0.93 & 1.00 & 0.96 & 0.97 & 1.00 & 0.98 & 0.85 & 1.00 & 0.92 & 0.84 & 0.92 & 0.88 \\
Pituitary tumor  & 0.92 & 0.98 & 0.95 & 1.00 & 0.98 & 0.99 & 0.94 & 0.97 & 0.95 & 0.94 & 0.98 & 0.96 \\
\hline
\end{tabular}
\label{tab1}
\end{table*}

\begin{table*}[htbp]
\caption{Model comparison across studies}
\begin{center}
\begin{tabular}{ c c c c c c} 
\hline
\textbf{Study no.} & \textbf{Models used} & \textbf{Accuracy} & \textbf{F1 Score} & \textbf{Precision} & \textbf{Recall} \\
\hline
\multirow{2}{*}{\textbf{\cite{b15}}}
             & Decision Tree & 0.96 & 0.9637 & 1.00 & 0.93\\
             & Naive Bayesian & 0.882 & 0.91 &0.91 &0.91 \\
\hline

\textbf{\cite{b16}} & Support Vector Machine & 0.971 & - & - & 0.919 \\
 \hline

\textbf{\cite{b17}} & MN-V2/CFO & 0.9732 & 0.8622 & 0.9768 & 0.8012 \\
\hline
\textbf{\cite{b18}} & ResNet50 & 0.9514 & 0.9515 & 0.9517 & 0.9514 \\ \hline
\multirow{4}{*}{\textbf{This study}} & VGG16 & 0.93 & 0.9325 & 0.9275 & 0.94 \\
                                     & EfficientNet &\textbf{ 0.98} & 0.9825 & 0.9825 & 0.9875\\
                                     & ResNet152V2 & 0.91 & 0.9050 & 0.9025 & 0.92 \\
                                     & ViT & 0.91 & 0.9025 & 0.90 & 0.9075\\
\hline
\end{tabular}
\label{tab3}
\label{tab2}
\end{center}
\end{table*}

Fig. \ref{fig5} illustrates the ROC curves of four deep learning models (VGG16, EfficientNet, ResNet152V2, and ViT) utilized for brain tumor classification. Models exhibiting AUCs approaching 1.00 demonstrate elevated accuracy. ViT and EfficientNet attain impeccable performance (AUC 1.00), although VGG16 and ResNet152V2, exhibiting AUCs near 1.00, demonstrate robust classification with minor discrepancies, indicating greater challenges in differentiating tumor kinds.

Fig. \ref{fig6} presents the confusion matrix, illustrating that EfficientNet shows exceptional classification performance across all tumor types, particularly in distinguishing glioma from non-tumor cases. The ViT and VGG16 also demonstrate strong performance, excelling particularly with meningioma tumors. ResNet152V2, however, struggles to differentiate glioma from the absence of a tumor, leading to the highest rate of misclassifications.
While all models perform well, EfficientNet stands out with the highest precision but ViT and VGG16 exhibit difficulties in glioma classification. Conversely, ResNet152V2 requires further tuning due to its challenges in correctly classifying glioma and no-tumor cases.

Table \ref{tab1} evaluates the efficacy of four models (VGG16, EfficientNet, ResNet152V2, and ViT) using F1 scores and accuracy in the classification of brain tumors. EfficientNet surpasses all models, achieving the greatest F1 scores and an overall accuracy of 0.98, exemplifying the optimal equilibrium between precision and recall. VGG16 exhibits commendable performance, particularly for meningiomas and pituitary tumors, achieving an accuracy of 0.93. ResNet152V2 and ViT have comparable overall accuracies (0.91); nevertheless, their performance is somewhat inconsistent, especially in the glioma and no tumor categories. The ViT demonstrates superior performance in the classification of pituitary tumors. 

In summary, EfficientNet is the most dependable and precise model, with VGG16 closely trailing behind. ResNet152V2 and ViT require enhancement for improved consistency. 

Table \ref{tab3} illustrates multiple brain tumor classification studies using machine learning and deep learning models, emphasizing this work's excellent results. The findings show that modern deep learning architectures outperform conventional methods and models. This study shows model performance improvements, with EfficientNet obtaining 0.98 accuracy. It shows its improved brain tumor classification, surpassing previous findings. With an accuracy of 0.93, VGG16 is reliable for various categorization jobs. ResNet152V2 and ViT score 0.91, showing good and consistent performance but slightly lower than EfficientNet. However, prior investigations yielded different results. Study \cite{b15} uses classical machine learning algorithms like Decision Trees and Naive Bayes. The Decision Tree performs well with structured data, but the Naive Bayesian classifier struggles with complex brain tumor data at 0.882. Support Vector Machines (SVMs) with different kernel functions achieve 0.97 accuracy in the study \cite{b16}. The work \cite{b17} employs MobileNetV2, a deep learning model that has been tuned using a unique metaheuristic known as the Contracted Fox Optimization Algorithm (MN-V2/CFO), to optimize its hyperparameters. This approach achieves a tumor detection accuracy of 97.32\%, outperforming the ResNet50 model described in \cite{b18}, which achieves an accuracy of 95.14\% on pre-processed MRI images. EfficientNet sets a new standard for brain tumor classification, showing that deep-learning networks can handle complex medical imaging tasks with precision.

EfficientNet demonstrates superior performance compared to traditional convolutional neural networks in the classification of brain tumors by utilizing compound scaling. This approach effectively balances depth, width, and resolution, thereby improving accuracy while preserving computational efficiency. The design is pretrained on extensive datasets, thereby facilitating improved feature extraction and generalization.

To conclude, the current study exhibits a distinct advancement compared to conventional and prior deep learning methods. EfficientNet establishes a new standard for the accuracy of brain tumor classification, demonstrating the capability of sophisticated deep-learning networks to manage intricate medical imaging tasks with exceptional precision.

\section{Conclusion}
The primary objective of the project is to categorize MRI images using advanced deep-learning models. The efficacy of various deep learning architectures, including ResNet152V2, VGG, ViT, and EfficientNet, is demonstrated. Recent improvements in MRI image denoising have shown the effectiveness of hybrid CNN models integrated with anisotropic diffusion filters. These advanced models are meticulously designed to extract critical features from MRI data. The comparative analysis has clarified the distinct advantages and limitations of each engineering discipline, facilitating informed decision-making in specific clinical situations. Furthermore, the adept execution of exchange learning has accelerated training and enhanced performance, while EfficientNet has demonstrated its ability to achieve high accuracy with a remarkable computing economy. The presented models demonstrate robustness against fluctuations in picture quality, patient demographics, and tumor kinds, indicating their suitability for diverse clinical settings. The models achieved an impressive total accuracy of 98\%, thereby validating their efficacy. The Contingency table, marked by a few false positives and false negatives, highlights the model's capability to accurately differentiate between tumor and non-tumor regions. This research has significantly advanced therapeutic image processing and provides a feasible method to better patient outcomes and refine clinical procedures in brain tumor detection.

\end{document}